\documentclass[11pt,a4paper]{article}
\usepackage[hyperref]{acl2020}
\usepackage{times}
\usepackage{latexsym}
\usepackage{subcaption}

\usepackage{graphicx}
\newcommand{\quotes}[1]{``#1"}
\usepackage{microtype}
\usepackage{amsmath, mathtools, amssymb}

\aclfinalcopy 
\def\aclpaperid{73} 

\title{How Effective is Incongruity? Implications for Code-mix Sarcasm Detection}

\author{Aditya Shah \thanks{* Work done while the author was an intern at IIT Indore} \\
   CSE department \\
 IIT Indore \\
 India \\
  \texttt{adishah3103@gmail.com} \\\And
 Chandresh Kumar Maurya \\
  CSE department \\
 IIT Indore \\
 India \\
  \texttt{chandresh@iiti.ac.in} \\}

\date{}

\begin{document}
\maketitle
\begin{abstract}
The presence of sarcasm in conversational systems and social media like chatbots, Facebook, Twitter, etc. poses several challenges for downstream NLP tasks. This is attributed to the fact that intended meaning of a sarcastic text is contrary to what is expressed. Further, the use of code-mix language to express sarcasm is increasing day by day.  Current NLP techniques for code-mix data have limited success due to the use of different lexicon, syntax, and scarcity of labeled corpora. To solve the joint problem of code-mixing and sarcasm detection, we propose the idea of capturing incongruity through sub-word level embeddings learned via fastText. Empirical results shows that our proposed model achieves F1-score on code-mix Hinglish dataset comparable to pre-trained multilingual models while training {\bf 10x} faster and using \emph{lower memory footprint}.
\end{abstract}

\section{Introduction}
Sarcasm is defined as a sharp remark whose intended meaning is different from what it looks like. For example, \quotes{{\it I am not insulting you. I am describing you.}} could mean that the speaker is insulting the audience, but the receiver does not get it. Sarcasm usually involves ambivalence (also known as \emph{incongruity} which means words/phrases having contradictory implications \cite{xiong2019sarcasm} and difficult to comprehend. Though English is used as a way to communicate and exchange messages, majority of the people still use the mother language to express themselves on social media \cite{danet2007multilingual}. According to one study \cite{hong2011language}, more than 50\% posts on Twitter are written in a language other than English. \emph{Code-switching} (also known as \emph{code-mixing}) is a writing style in which the author uses words from different languages either in the same sentence (called \emph{intra--sentential}) or different sentences (called \emph{inter--sentential}) switching. An example of code-switching is:\emph{\quotes{he said kal karte hai kaam"}} (Gloss: he said tomorrow we'll do the work). The studies such as \cite{vizcaino2011humor,siegel1995get} show that people use code-switch language when trying to convey comicality, satire or humor. Motivated by previous studies, we plan to detect sarcasm in code-mix languages. Though many studies exist to detect sarcasm in unimodal and multimodal data \cite{joshi2015harnessing, joshi2017automatic,carvalho2009clues,xiong2019sarcasm,cai2019multi}, methods to detect sarcasm in code-mix data are limited and have not been explored much  \cite{bansal2020code,aggarwal2020did,swami2018corpus}. It is due to several challenges such as \emph{ambiguous words, variable lexical representation, word-level code-mixing, reduplication}, and \emph{word-order}. 

To solve some of these issues, we present a deep-learning based architecture to capture incongruity in code-mix data. Our proposed model achieves competitive performance as compared to pre-trained multilingual models (fine-tuned on the code-mix sarcasm detection task) with significantly \emph{fewer parameters and faster training time}. Our contributions are as follows:
\begin{itemize}
    \item Propose a deep learning based architecture along with sub-word level features to capture incongruity for sarcasm detection.
    \item Evaluate the performance of the proposed model on the Hindi-English (Hinglish) code-mix Twitter data that we collected. We further analyze existing multilingual models on the same.  Our code+data will be available on \footnote{https://github.com/likemycode/codemix}.

    \item We will release the benchmark sarcasm dataset for Hinglish language to facilitate further research on code-mix NLP.
\end{itemize}
\section{Related Works}
\subsection{Learning Representation for Code-Mix Data}
Models developed for multilingual representation learning have been explored for code-mix data representation by several authors \cite{winata2021multilingual,khanuja2020gluecos,aguilar2020lince,winata2018bilingual}. Character-level representations have been utilized
to address the out-of-vocabulary (OOV) issue in code-switch text \cite{winata2018bilingual}, hand-crafted features were used in \cite{aguilar2019multi} for handling low-resource scenarios. Fine-tuning multilingual models like mBERT has shown to yield good results for various NLP tasks like Named-entity recognition (NER), part-of-speech (POS) tagging, etc., in \cite{khanuja2020gluecos}, and surprisingly outperforms cross-lingual embeddings. Meta-embedding and hierarchical meta-embeddings have been found to be useful for closely-related language pairs in code-mix data \cite{winata2021multilingual} and usually outperform the mBERT \cite{khanuja2020gluecos}.
Char2Subword model proposed by \cite{aguilar2020char2subword} builds representations from characters out of the subword vocabulary, and uses them to replace subwords in code-mix text \cite{winata2018bilingual}, hand-crafted features were used in \cite{aguilar2019multi} for handling low-resource scenario.
A centralized benchmark for Linguistic Code-switching Evaluation (LinCE)  is released in \cite{aguilar2020lince,khanuja2020gluecos}. Both of these works present results on several NLP tasks but \emph{sarcasm detection}.
\subsection{Sarcasm Detection in Code-Mix Data}
There exists only a few works targeted towards sarcasm detection in code-mix data. In \cite{aggarwal2020did}, the author experiments with FastText \cite{joulin2016bag} and Word2Vec embeddings on two kinds of data: (1) Hinglish (Hindi-Eng) tweets, and (2) Hinglish+English tweets. They find that that Hinglish+English combination produces better results and achieves best F1 score of 79.4\%. Various hand-crafted features, such as char n-grams, word n-grams etc., combined  with random-forest/SVM are explored in \cite{swami2018corpus} for sarcasm detection in code-mix, data and achieve F1-score of 78.4\%. However, the dataset used  is highly imbalanced with just 10\% of sarcastic tweets and rest non-sarcastic. In such a scenario, the model might be biased towards predicting non-sarcastic tweets, and hence the evaluation results are quite skewed.  Along similar lines, different switching features are used to form feature-vector and fed into a hierarchical attention network in \cite{bansal2020code}. They find that switching feature is a good indicator for irony/sarcasm/hate speech detection. However, none of these works handles \emph{incongruity} explicitly or implicitly which has shown to achieve impressive results in sarcasm detection \cite{xiong2019sarcasm}.
\section{Model Architecture}

\begin{figure}
    \centering
    \includegraphics[width=8cm, height=6cm]{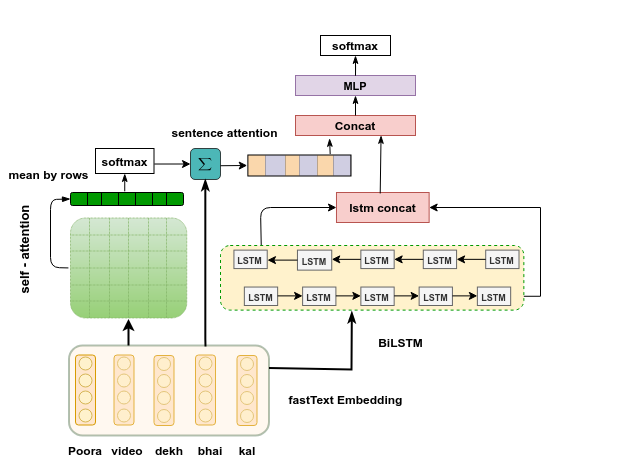}
    \caption{The model architecture}
    \label{fig:model}
\end{figure}

Code-mix language contains noisy words mixed with different languages and this might lead to out-of-vocabulary $<OOV>$ tokens. So, we use FastText  skipgram \cite{bojanowski2016enriching, Grave2018LearningWV} for learning \emph{subword level representation} from the code-mix data. We hypothesise that subword level representation is able to handle \emph{ambiguous words}, \emph{variable lexical representation}, and \emph{word-level code-mixing}. For example, the ambiguous word \quotes{to} may be present in both Hindi and English language. Learning subword representation alleviates the problem of encoding the word \quotes{to} differently for both the languages. Further, variable length words like \{\quotes{gharr}, \quotes{gharrr}, \quotes{gharrrr} will be split into tokens \{\quotes{gha},\quotes{har}, \quotes{arr},\quotes{rrr} \} and uniquely represented using only these subwords tokens. Code-mix words like \quotes{chapless} (Mix of Bengali \quotes{chap} and English \quotes{less}) are also represented via sub-words \quotes{chap} and \quotes{less}. The proposed model architecture is shown in Fig. \ref{fig:model}.

Each sentence $s$ is represented by its embedding $E=[e_1^T,e_2^T,\ldots,e_n^T ]$, where $e_i \in R^d$ is the embedding vector and $n$ is the length of the sentence. 
Inspired by the work of \cite{xiong2019sarcasm}, we propose the use of self-matching network in order to capture \emph{incongruity} within the code-mix sentence. Specifically, for word-embedding pairs $(e_i,e_j)$, we first calculate the joint feature vector $m_{i,j}$ via 
\begin{equation}
    m_{i,j}=GELU(e_i^T\cdot M_{i,j}\cdot e_j)
\end{equation}
where $M_{i,j} \in R^{d \times d}$ is the weight parameter matrix (learnable) and \emph{GELU} \cite{gelu} is Gaussian Error Linear Unit activation function. Instead of \emph{tanh} as used in \cite{xiong2019sarcasm}, we use \emph{GELU}  as it provides well-defined gradients in the negative region. Compared to RELU, since GELU is differentiable for all input values so it is widely used in state-of-the-art NLP architectures. Our findings suggest that GELU activation yields better results for attention based mechanisms. Note that the above formulation is a form of \emph{bilinear similarity} popularly used in \emph{metric learning}.  
To calculate the attention score $\alpha_i, i\in (1,2,\ldots,n)$ for each word, we take the mean of each row (contrary to \emph{max} of rows as in \cite{xiong2019sarcasm}) and apply the softmax for normalization.
\begin{equation}
    \alpha_i = softmax(\mu(m_{1,i}), \mu(m_{2,i}),\ldots, \mu(m_{n,i}))
\end{equation}
where $\mu()$ is the mean function, $m_{1,i}$ captures the incongruity of the first word with every other $i^{th}$ word in the input text. Using the mean of all attention scores considers all the incongruous words present in the sentence for the computation. This helps the model to attend and learn from much larger span of incongruity. These attention scores denote how much weight should be assigned to each incongruous word. Next we calculate the weighted sentence attention vector $v$ by:
\begin{equation}
    v = \mathbf{\alpha}^TE
\end{equation}
Self-attention approach though captures the \emph{incongruity} in the sentence, it misses the sentence's compositionality which is essential for sarcasm detection as suggested in \cite{tay2018reasoning}. Therefore, sentence embedding is passed to the BiLSTM encoder \cite{graves2013speech} and the hidden states of the forward LSTM and backward LSTM are concatenated as 
\begin{equation}
    h_i=[\overrightarrow{LSTM}(e_i),\overleftarrow{LSTM}(e_i)], \forall i \in (1,2,\ldots,n)
\end{equation}
The output of the BiLSTM is concatenated with the sentence attention vector $v$ and passed through the MLP layers with dropout. Finally, we pass it through softmax to predict the distribution over the binary labels (sarcasm vs non-sarcasm).
\begin{equation}
    \hat{y} = softmax(MLP([v,h_n]))
\end{equation}
where $h_n$ is the hidden state corresponding to the last word in the forward and backward LSTM.
\section{Experiments}
\subsection{The Dataset}
The code-mix dataset used by \cite{aggarwal2020did} is highly imbalanced with just 10\% of sarcastic tweets and rest
non-sarcastic. So, we create a dataset using TweetScraper built on top of scrapy \footnote{https://github.com/jonbakerfish/TweetScraper} to extract code-mix hindi-english tweets. 
We pass search tags like \#sarcasm, \#humor, \#bollywood, \#cricket, etc., combined with most commonly used code-mix Hindi words as query. 
All the tweets with hashtags like \#sarcasm, \#sarcastic, \#irony, \#humor etc. are treated as positive. Non sarcastic tweets are extracted using general hashtags like \#politics, \#food, \#movie, etc. The balanced dataset comprises of 166K tweets. We preprocess and clean the data by removing urls, hashtags, mentions, and punctuation in the data. 

\begin{table*}
\centering
\small
\caption{Comparative evaluation of the proposed approach.}
\label{results}
\begin{tabular}{l|c|c|c|c|c|c|c}
\hline \textbf{Model} & \textbf{Recall} & \textbf{Prec.} & \textbf{Acc.} & \textbf{F1} & \textbf{Params} & \textbf{GPU} & \textbf{Train time} \\ \hline
Attn. BiLSTM     & 77.34     & 81.24    & 80.21  & 79.34    & 21M    & 68 MB           &  0.8Hr    \\
XLM-RoBERTa & 86.17 & 91.48 & 89.04 & 88.75  & 278M   & 575 MB          &8 Hr\\
mBERT & 83.20 & \textbf{94.55} & \textbf{89.17} & 88.51  & 167M   & 483 MB          &7 Hr \\ \hline
SelfNet (Ours) & \textbf{88.12 } & 88.25 & 89.04 & {\bf 88.89} & 35M    & 80 MB           & 1 Hr   \\
\hline
\end{tabular}
\end{table*}

\begin{figure*}
\centering
	\begin{subfigure}{1in}
		\centering
		\includegraphics[width=5cm, height=4cm]{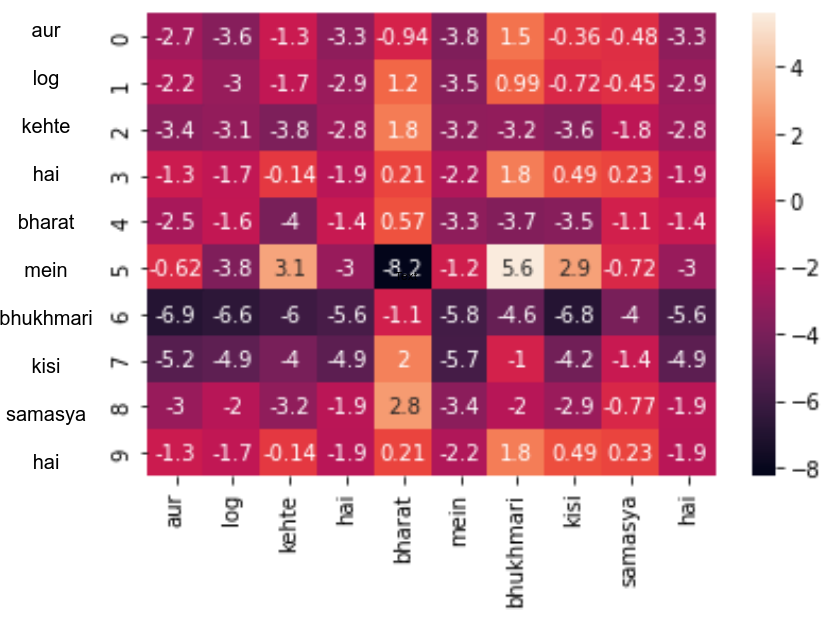}
		\caption{Sarcastic}\label{fig:1a}		
	\end{subfigure}
	\hspace{8em}
	\begin{subfigure}{1in}
		\centering
		 \includegraphics[width=5cm, height=4cm]{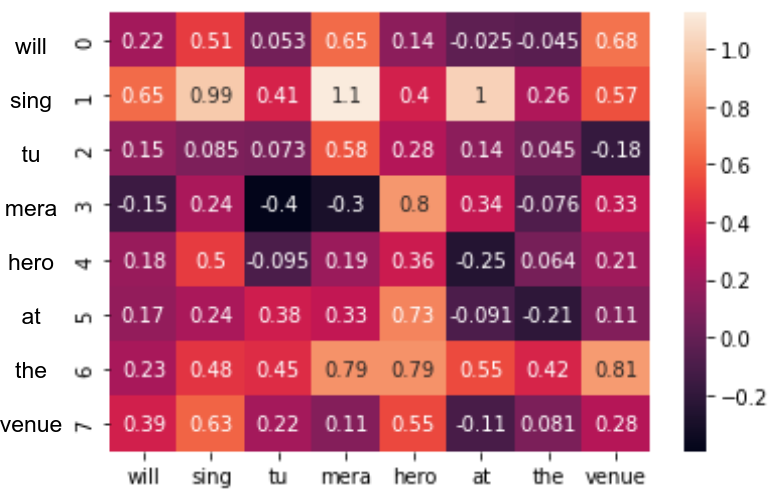}
		\caption{Non Sarcastic }\label{fig:1b}
	\end{subfigure}
	\caption{Incongruity  visualization via attention matrix }\label{maps}
\end{figure*}

\subsection{Baselines}
The following baselines are used for comparison. (a) {\bf  Attention BiLSTM} \cite{aggarwal2020did}: The text features are extracted using word2vec and FastText which is fed to Series CNN, Parallel CNN, LSTM, Bi-LSTM and Attention Bi-LSTM, (b) {\bf Multilingual Models}: To showcase the competitiveness of the proposed approach, we also compare with the state-of-the-art multilingual models like XLM-RoBERTa\footnote{https://tinyurl.com/ydseww9d} and mBERT\footnote{https://tinyurl.com/2dafn48n} from Huggingface library \cite{huggingface}. Specifically, we first fine-tune these models on the preprocessed code-mix corpus for mask language modeling task. Next, we use trained model by attaching a dense layer on top of it for detecting sarcasm in the code-mix tweets. 
 
\subsection{Experimental setup}
For all the experiments, we use a train/valid/test split of 65:15:20. Categorical cross-entropy loss is minimized using adam optimizer for 15 epochs and learning rate of 5e-4 with step wise learning rate scheduler. FastText embedding size is 100 and the number of hidden units in BiLSTM and MLP layers are 256. We apply dropout of 0.4 along with gradient clipping of 0.3.

\subsection{Results}
The comparative evaluation results are shown in Table \ref{results}. We can see that the proposed approach achieves better F1 score than the baselines on the Hinglish code-mix data. In particular, our approach achieves around 10 points more F1 score than the Attn. BiLSTM of \cite{aggarwal2020did}. Additionally, it achieves slightly better F1 score than the pre-trained multilingual models.

Further, we also provide a comparison among all the 4 models in terms of the trainable parameters, GPU memory and training time in Table \ref{results}. As it can be seen that multilingual models require much larger memory and use almost 10x more parameters than our approach. Also, it is worth noting that the dataset size used to train our model is significantly less than the dataset size used to train multilingual models. Based on the comparison, we observe that our proposed model achieves a good balance between performance and model size for code-mix sarcasm detection. Figure \ref{maps} illustrates the output raw attention matrix $P$ obtained before applying activation to visualize incongruous words. 

As we can see from the Figure \ref{fig:1a}, the words \emph{bharat, mein, and bhukhmari} hold the highest incongruity (negative values). These 3 words define the semantics of the sentence and our model correctly attends to those words while finding the incongruity. Similarly for Figure \ref{fig:1b}, there's no such incongruity present in the text. Thus the model does not assign high negative scores to this matrix.

\subsection{Ablation Study}
To evaluate the effectiveness of the network, we conduct an ablation study on the proposed architecture. This is summarized in Table \ref{ablation}. We test the self matching network proposed by \cite{xiong2019sarcasm} which is referred to as Self Matching Net. Next, we replace biLSTM in our model with XLM-RoBERTa and mBert. The resulting models are denoted by \quotes{with XLM-RoBERTa} and \quotes{with mBert} respectively. The original Self Matching network does not perform so well on code-mix data as it only considers the most incongruous word pairs for prediction. Using mean operation helps to capture all the incongruous words which results in performance gain. Next when we replace BiLSTM with the multilingual models, the resulting approaches do not perform better than the proposed model. Although these models are trained on huge multilingual corpus, our study suggests that we can capture nuances of code-mix language using self-attention and simpler models like BiLSTM in a better way.

\begin{table}[h]
\centering
\small
\caption{ Ablation Study}
\label{ablation}
\begin{tabular}{ l|c|c|c|c}
\hline \textbf{Models} & \textbf{Recall} & \textbf{Prec.} & \textbf{Acc.} & \textbf{F1} \\ \hline
Self Matching Net & 81.68 & 81.55 & 81.7 & 81.68\\
with XLM-RobertA & 87.71 & 87.73 & 87.81 & 87.81  \\
with mBERT & 86.94 & 86.72 & 86.85 & 86.94 \\  \hline
SelfNet (Ours) & \textbf{88.12} & \textbf{88.25} & \textbf{89.04} & \textbf{88.89} \\
\hline
\end{tabular}
\end{table}

\section{Conclusion \& Future work}
In the present work, we propose the significance of incongruity in order to capture sarcasm in code-mix data. Our model effectively captures incongruity through FastText sub-word embeddings to detect sarcasm in the text. Empirical results on code-mix sarcasm data show that our approach performs satisfactorily compared to the multilingual models while saving memory footprint and training time.  In future, we plan to work on a generalized model for other code-mix NLP tasks (NLI, NER, POS, QA etc) as well as test other code-mix languages like English - Spanish, English - Tamil, English - French etc.

\bibliography{anthology,acl2020}
\bibliographystyle{acl_natbib}

\end{document}